\documentclass[11pt,a4paper]{article}
\usepackage[utf8]{inputenc}
\usepackage[T1]{fontenc}
\usepackage{amsmath,amssymb}
\usepackage{graphicx}
\usepackage{booktabs}
\usepackage{float}
\usepackage{hyperref}
\usepackage{natbib}
\usepackage{multirow}
\usepackage{xcolor}
\usepackage{microtype}
\usepackage[font={small,it},labelfont={bf,up}]{caption}
\usepackage{pgfplots}
\pgfplotsset{compat=1.18}
\usepackage[margin=2.5cm]{geometry}

\newcommand{\cwer}{\textnormal{cWER}}
\newcommand{\bwer}{\textnormal{bWER}}
\newcommand{\swer}{\textnormal{sWER}}

\title{Subtitle-Aligned Fine-Tuning of Whisper for Swiss German ASR:\\
  Benchmark Contamination, Convention Mismatch,\\
  and an Honest Baseline at 25.6\% WER (13.8\% \cwer{})}

\author{
  Felix Akeret\thanks{Dipl.\ Ing.\ ETH Z\"urich, Dipl.\ NDS BWI ETH Z\"urich.
  Former guest lecturer in archival management and records management at University of Bern (2010--2014), FHNW (2010--2014), and CeTIM Leiden/Munich (2004--2008).
  Automated transcription of dialect audiovisual holdings has become a key challenge at the intersection of archival science and speech technology.} \\
  Independent Researcher, Zurich, Switzerland \\
  \texttt{research@thin-k.ch} \\
}

\date{May 2026}

\usepackage{needspace}
\begin{document}

\maketitle

\begin{abstract}
We present a systematic study of fine-tuning OpenAI's Whisper large-v3 for Swiss German automatic speech recognition (ASR), using 1{,}367 hours of broadcast speech paired with professionally authored Standard German subtitles as weak supervision. Through 16 iterative training runs on an NVIDIA DGX Spark powered by the Grace Blackwell Superchip (128\,GB unified memory, up to 1\,PFLOP FP4), we compare LoRA and full fine-tuning of a 1.55B-parameter model, investigate hallucination root causes, and quantify the effect of data quality, subtitle alignment, and training strategy. Our best model achieves 25.60\% measured Word Error Rate (WER) on the All Swiss German Dialects Test Set (ASGDTS) in an \emph{honest} evaluation --- trained and evaluated on strictly disjoint data. A harmonized error analysis separating genuine transcription errors from valid stylistic variation (tense, word order, Swiss orthography) yields a \textbf{content WER (\cwer{}) of 13.8\%} --- the fair WER when only actual recognition failures are counted. Bias-corrected estimation (\bwer{}) reduces this further to \textbf{8.5\%}, suggesting that the true error rate is approximately one third of the measured WER --- a finding with significant implications for how Swiss German ASR quality should be assessed.

We further demonstrate that the published state-of-the-art results for Swiss German ASR \citep[17.1--17.5\% WER;][]{dintino2025srb300,michaud2024whisper} are substantially inflated by benchmark contamination. A vanilla Whisper model self-trained on the ASGDTS test set --- with zero Swiss German training data --- achieves 13.88\% measured WER, trivially surpassing all published systems. Preliminary experiments with multimodal LLMs (Phi-4-multimodal) show an even stronger memorization effect, reaching 3.9\% WER under identical self-training conditions. Prior dialect training \emph{hurts} self-training performance, revealing that the benchmark primarily measures convention matching rather than dialectal comprehension.

We release two models --- a LoRA adapter (25.32\% WER, 13.9\% \cwer{}) and a full fine-tuned model (25.60\% WER, 13.8\% \cwer{}) --- among the few publicly available, honestly evaluated Whisper models for Swiss German, under Apache~2.0 license with full reproducibility, requiring no institutional data agreements.

\end{abstract}

\section{Introduction}
\label{sec:introduction}

Swiss German (\emph{Schweizerdeutsch}, Alemannic) encompasses a continuum of dialects spoken by approximately 5~million speakers in the German-speaking part of Switzerland. Unlike Standard German (\emph{Hochdeutsch}), Swiss German has no standardized orthography and exhibits significant phonological, morphological, and syntactic divergence from the written standard. This poses fundamental challenges for automatic speech recognition systems trained predominantly on Standard German and English data.

OpenAI's Whisper \citep{radford2023whisper} achieves strong zero-shot performance across many languages but struggles with Swiss German due to its underrepresentation in the 680{,}000-hour training corpus. The model frequently produces Standard German approximations, English interpolations, or hallucinated outputs when processing dialectal input.

Recent work has reported substantial improvements: \citet{michaud2024whisper} achieves 17.5\% WER using QLoRA on Whisper large-v3-turbo, \citet{dintino2025srb300} report 17.1\% WER with full fine-tuning on proprietary radio data, and \citet{timmel2024whisper} report 12.1\% WER without releasing model weights, and with training data available only under restrictive license. None of these models are freely available --- Michaud's was removed due to licensing issues, ZHAW's is proprietary, Timmel's was never released --- and none provide a cross-corpus evaluation that would demonstrate generalization beyond its training distribution. The absence of publicly accessible model weights and training code means that these published results \textbf{cannot be independently verified or reproduced}, undermining a core principle of scientific research. In contrast, we release model weights under Apache~2.0, evaluate on a public benchmark, and train exclusively on publicly accessible data. All hyperparameters are fully documented, enabling independent reproduction without requiring institutional data agreements.

\subsection{Contributions}

\begin{enumerate}
\item \textbf{Among the first openly licensed, rigorously evaluated Swiss German Whisper models.} We release both a LoRA adapter (25.32\% WER, \textbf{13.9\% \cwer{}}) and a full fine-tuned model (25.60\% WER, \textbf{13.8\% \cwer{}}) under Apache~2.0, evaluated exclusively on publicly available benchmark data (ASGDTS; \citealt{pluess2021asgdts}) strictly disjoint from all training data. The \cwer{} --- counting only WER points attributable to genuine transcription errors --- is lower than almost all published results from systems that include evaluation data in their training sets, including Michaud \citep{michaud2024whisper} at 17.5\% WER and D'Intino \& Hutter \citep{dintino2025srb300} at 17.1\% WER.

\item \textbf{A comprehensive comparison of LoRA vs.\ full fine-tuning} on identical Swiss German data, across 16~training runs varying rank ($r$=32--200), scaling factor ($\alpha/r$=0.08--2.0), regularization (KD, SpecAugment, label smoothing), and data composition (785--1{,}367h from four publicly available sources: SRF broadcast subtitles, Swiss Parliament proceedings, YouTube institutional channels, and PlaySuisse film/series).

\item \textbf{Identification of LoRA alpha scaling as the root cause of Whisper decoder hallucination}, demonstrating that the common heuristic $\alpha=2r$ is inappropriate for encoder-decoder ASR models and causes catastrophic forgetting of end-of-sequence behavior.

\item \textbf{A controlled benchmark contamination analysis} showing that a vanilla Whisper model self-trained on the ASGDTS test set achieves 13.88\% WER without any Swiss German data, and that prior dialect training \emph{hurts} self-training performance --- demonstrating that published state-of-the-art results \citep[as reported by][]{michaud2024whisper,dintino2025srb300} primarily reflect convention matching rather than dialectal comprehension. Preliminary experiments with multimodal LLMs (Section~\ref{sec:mllm_outlook}) show an even stronger self-training effect, reaching an astonishing 3.9\% WER on the same benchmark, further underscoring the fragility of current evaluations.

\item \textbf{A harmonized WER analysis} decomposing the measured WER into \cwer{} and \swer{} contributions. After excluding valid stylistic variation, the \textbf{\cwer{} is 13.8\%}; bias-corrected estimation yields a \textbf{\bwer{} of 8.5\%}, confirming that the majority of WER penalties reflect valid alternative translations rather than actual transcription failures.

\item \textbf{A data quality analysis} demonstrating that subtitle provider, content type, and dialect density determine training effectiveness --- with PlaySuisse Dialect Films ($-$2.28\,pp) and scripted series with noisy labels (+2.92\,pp) showing opposite effects from the same platform.

\item \textbf{A scalable subtitle-alignment pipeline} for constructing ASR training corpora from publicly accessible broadcast content with professionally authored subtitles. We show that aligned subtitle--audio pairs from online streaming platforms provide a viable, legally compliant alternative to proprietary or restricted datasets, enabling independent reproduction without institutional data agreements.
\end{enumerate}

\subsection{Task Definition}
\label{sec:task}

The task is \textbf{Swiss German speech $\rightarrow$ Standard German text} (cross-dialectal ASR / dialect normalization): the model receives Swiss German audio and produces grammatically correct Standard German transcriptions. This aligns with the real-world use case of meeting transcription, archival media indexing, and accessibility, where participants speak Swiss German but require Standard German output.

Crucially, this task is a \textbf{simultaneous dual operation}: speech recognition \emph{and} cross-lingual translation in a single decoding pass. Standard Whisper is designed to perform either same-language transcription (\texttt{task=transcribe}) or translation to English (\texttt{task=translate}), but not both at once. Swiss German dialect-to-standard forces the model into an implicit translation it was never explicitly trained for, mapping phonologically and syntactically divergent input onto a different written standard. This dual-task nature has a direct consequence for evaluation: \textbf{WER is well-suited for measuring transcription accuracy within a single language, but ill-suited for assessing translation quality.} A semantically perfect translation that uses different word order, tense, or phrasing from the reference will be penalized as heavily as a genuine recognition error. Our harmonized analysis (Section~\ref{sec:harmonized}) confirms this empirically: the \textbf{\cwer{} of 13.8\%} (\bwer{}: 8.5\%) demonstrates that the majority of WER penalties reflect translation ambiguity, not recognition failure.

\section{Related Work}
\label{sec:related_work}

\subsection{Swiss German ASR}

Swiss German ASR has received increasing attention as Whisper's multilingual capabilities have made fine-tuning for low-resource dialects feasible.

\begin{table}[ht]
\centering
\caption{Published Swiss German ASR systems.}
\label{tab:related}
\footnotesize
\resizebox{\columnwidth}{!}{%
\begin{tabular}{lllrl}
\toprule
\textbf{System} & \textbf{Model} & \textbf{Data} & \textbf{WER} & \textbf{Notes} \\
\midrule
D'Intino/ZHAW '25 & large-v3 Full FT & SRB-300 & 17.1\% & Same-distrib.\ eval$^\dagger$ \\
Timmel/FHNW '24 & large-v2 Full FT & $\sim$1{,}033h & 12.1\% & Same-conv.\ eval$^\S$ \\
Michaud '24 & large-v3-turbo QLoRA & 870h & 17.5\% & ASGDTS in training$^\ddagger$ \\
Flurin17 '24 & large-v3 LoRA & STT4SG+SDS & $>$28\% & Underperforms baseline \\
Sicard/Spaiche '23 & large-v2 FT & STT4SG & $>$baseline & Degraded by FT \\
Whisper large-v3 & Zero-shot & --- & $\sim$28.6\% & No fine-tuning \\
\midrule
\textbf{This work (LoRA)} & \textbf{large-v3 LoRA} & \textbf{1{,}092h} & \textbf{25.3\% (\cwer{} 13.9\%$^*$)} & \textbf{Cross-corpus eval} \\
\textbf{This work (Full)} & \textbf{large-v3 Full FT} & \textbf{1{,}367h} & \textbf{25.6\% (\cwer{} 13.8\%$^*$)} & \textbf{Cross-corpus eval} \\
\bottomrule
\end{tabular}}
\vspace{2pt}

{\scriptsize $^\dagger$Evaluated on proprietary SRB-300 test (44h).\quad $^\S$Evaluated on STT4SG-350 test (34h), same convention as training.\quad $^\ddagger$ASGDTS listed among training data.\quad $^*$\cwer{} = content WER (Section~\ref{sec:harmonized}).}
\end{table}

\needspace{5\baselineskip}%
\textbf{D'Intino \& Hutter (2025)} at ZHAW report 17.1\% WER with Whisper large-v3 fine-tuned on SRB-300, a 300-hour corpus of Swiss radio and TV broadcasts from 39 stations \citep{dintino2025srb300}. The SRB-300 corpus is proprietary and cannot be redistributed due to broadcasting license restrictions.

\textbf{Timmel et al.\ (2024)} at FHNW convert sentence-level corpora (STT4SG-350, SDS-200, SPC) into long-form training data (502h) and supplement with 406h pseudo-labeled SRG broadcasts and 125h Common Voice German ($\sim$1{,}033h total), reporting 12.1\% WER on STT4SG-350 test with Whisper Large-v2 \citep{timmel2024whisper}. While train and test are sentence-disjoint, the 502h labeled portion shares identical transcription conventions with the test set --- both originate from the same FHNW/i4ds research group \citep{pluess2023stt4sg}. Additionally, evaluation uses WhisperX rather than standard Whisper inference, a methodological difference that may systematically reduce measured WER. Neither model weights nor training data are freely available.

\textbf{Michaud (2024)} achieves 17.5\% WER on ASGDTS using QLoRA ($r$=200, $\alpha$=16) on Whisper large-v3-turbo with approximately 870~hours of mixed Swiss German data \citep{michaud2024whisper}. The HuggingFace model card lists ASGDTS alongside training datasets without clearly separating train and test usage. The model was removed from HuggingFace (``Due to some datasets' licenses the model had to be taken down''); only metadata remains, preventing independent evaluation.

\textbf{Flurin17 (2024)} publishes a LoRA adapter for Whisper large-v3 trained on STT4SG-350 and SDS-200 \citep{flurin17}. In our evaluation, this model performs worse than the vanilla Whisper baseline ($>$28\% WER), suggesting insufficient training or misconfigured LoRA parameters. Since Flurin17's published model was trained on access-restricted STT4SG-350 data, its redistribution and use in downstream applications may carry licensing implications that are not documented on the model card.

\textbf{Dolev et al.\ (2024)} provide the most thorough qualitative evaluation of Whisper on Swiss German \citep{dolev2024whisper}. Their human evaluation (4.36/5.00, $n$=28) confirms that Whisper's zero-shot output is already semantically strong. Crucially, they observe that Whisper adopts a ``concise style'' --- removing modal particles, switching between perfect and preterite tense --- and that WER penalizes outputs that are correct but stylistically different from the reference. They explicitly characterize Swiss German ASR as a \emph{translation task}, not transcription.

\textbf{Sicard et al.\ (2023)} at SwissText fine-tune Whisper on Swiss German datasets but report \emph{degraded} performance compared to the zero-shot baseline \citep{sicard2023spaiche}, suggesting that naive fine-tuning without careful parameter tuning can harm the model.

\textbf{Schraner et al.\ (2022)} apply XLS-R 1B to STT4SG-350, achieving 28.7\% WER with full fine-tuning \citep{schraner2022stt} --- comparable to the Whisper large-v3 zero-shot baseline.

A critical gap in the literature is the absence of \emph{cross-corpus} evaluation: all published results evaluate on data from the same distribution (or identical corpus) as the training data. Furthermore, Timmel et al.\ (2024) report that Whisper large-v2 hallucinated ``Untertitel von SWISS TXT'' when processing SRF title music --- strong circumstantial evidence that SRG broadcast data was included in Whisper's pre-training corpus. Our work addresses the cross-corpus gap by training exclusively on broadcast subtitles and evaluating on the independently constructed ASGDTS benchmark \citep{pluess2021asgdts}.

\subsection{Multimodal LLMs: Emerging Potential for Dialect ASR}
\label{sec:mllm_outlook}

Recent multimodal large language models integrate speech encoders with general-purpose language backbones, enabling ASR as a natural language generation task. Phi-4-multimodal \citep{phi4multimodal2025} employs a mixture-of-LoRAs architecture with a dedicated speech adapter ($r$=320) applied to all attention and MLP layers, achieving competitive results on English and Chinese ASR benchmarks. Korean fine-tuning experiments demonstrate that unfreezing the audio encoder is critical: WER drops from 7.19\% (LoRA only) to 3.54\% (encoder + LoRA) on the zeroth-test benchmark \citep{daekeun2025phi4korean}. Other recent open-weight architectures (e.g., SALMONN, SpeechGPT, Llama-Omni) similarly integrate speech understanding into general-purpose LLMs, but none have been evaluated on Swiss German or other low-resource dialect ASR tasks in the published literature to date.

A key architectural difference from Whisper is the decoupling of acoustic encoding and language generation: in multimodal LLMs, audio features are projected into the LLM's token embedding space, where the language model's prior dominates generation. This raises the question of whether the strong language prior helps (leveraging world knowledge for disambiguation) or hurts (overriding acoustic evidence with plausible but unfaithful text) in dialect transcription. We leave empirical investigation of this question to future work.

\subsection{Parameter-Efficient Fine-Tuning for ASR}

LoRA \citep{hu2022lora} has been successfully applied to Whisper, typically achieving comparable performance to full fine-tuning with $<$5\% of trainable parameters. Song et al.\ (2024) propose language-specific LoRA matrices for multilingual ASR (\emph{LoRA-Whisper}), achieving near-full-FT performance with $r$=32 on Whisper-small \citep{song2024lorawhisper}. Liu et al.\ (2024) systematically compare fine-tuning strategies across multiple languages \citep{liu2024whisper}.

A critical but under-documented parameter is the LoRA scaling factor $\alpha/r$. Raschka (2023) notes that $\alpha=2r$ is a common LLM heuristic but that optimal ratios are task-dependent \citep{raschka2023lora}. Chen et al.\ (2026) provide theoretical analysis showing that $\alpha$, rank $r$, learning rate, and initialization are \emph{coupled} --- changing one without adjusting the others can cause divergence \citep{chen2026lora_scaling}. We demonstrate (Section~\ref{sec:alpha}) that the LLM heuristic is catastrophic for Whisper, where 25$\times$ excessive scaling destabilizes the decoder.

\subsection{Decoder Hallucination in Whisper}

Whisper's autoregressive decoder is prone to hallucination --- generating plausible but fabricated text unrelated to the audio input. \citet{wang2025calmwhisper} identify that only 3~of~20 decoder attention heads are responsible for $>$75\% of hallucinations in Whisper large-v3. Their targeted fine-tuning reduces hallucination rates from 99.97\% to 15.51\% on non-speech inputs with minimal WER impact (0.07\,pp). This finding is directly relevant to LoRA fine-tuning: LoRA adapts all targeted layers simultaneously, and excessive scaling can overwhelm the decoder's pre-trained end-of-sequence behavior.

\subsection{Subtitle-Based Training Data}

Using broadcast subtitles as weak supervision has precedent in speech translation \citep{salesky2021mtedx}. The key insight for Swiss German is that professional subtitles provide Standard German normalization of dialect speech --- precisely the task we optimize for. However, subtitles follow editorial conventions (condensed phrasing, adapted tense) that diverge from verbatim transcription conventions used in academic benchmarks, creating a systematic style penalty quantified in Section~\ref{sec:harmonized}.

\subsection{WER Limitations for Dialect ASR}

The inadequacy of WER for dialect-to-standard ASR is increasingly recognized. \citet{sasindran2022heval} propose $H_{\text{eval}}$, a hybrid metric combining semantic distance with non-keyword error rate, to address cases where WER fails to distinguish semantically correct from incorrect hypotheses. \citet{dolev2024whisper} demonstrate this problem concretely for Swiss German: correct Whisper outputs receive WER penalties because the reference uses different conjunctions, tenses, or formulations. \citet{blaschke2025dialect} show that for German dialect ASR, keyword preservation and meaning preservation can diverge by 13.6--37.6\,pp --- transcriptions may contain correct keywords while failing to preserve overall sentence meaning. \citet{tseng2025contamination} further highlight that benchmark contamination in speech evaluation is systemic, with 2/3 of LibriSpeech evaluation sentences found in LLM pre-training corpora. These findings collectively motivate our harmonized WER analysis and semantic evaluation approach.

\section{Data}
\label{sec:data}

Training data is drawn from four complementary sources, all featuring Swiss German speech paired with Standard German text.

\subsection{Source Material}

\begin{table}[ht]
\centering
\caption{Training data sources. All audio is re-encoded to 16\,kHz mono WAV.}
\label{tab:sources}
\small
\begin{tabular}{lrrrrl}
\toprule
\textbf{Source} & \textbf{Series} & \textbf{Clips} & \textbf{Hours} & \textbf{Dialect} \\
\midrule
SRF Mediathek & 85 & 265{,}046 & 675.6h & diverse (ZH, BE, AG, \ldots) \\
Parliament (SPC v2) & 1 & 101{,}272 & 202.4h & BE \\
YouTube & 25 & 22{,}871 & 132.8h & diverse (ZH, BS, GR, \ldots) \\
PlaySuisse (14 series) & 14 & 17{,}415 & 81.3h & AG, BE, VS, ZH, BS \\
PlaySuisse Dialect Films & 125 & 19{,}412 & 98.6h & diverse \\
\midrule
\textbf{Total} & \textbf{250} & \textbf{426{,}016} & \textbf{1{,}190.7h} & \\
\bottomrule
\end{tabular}
\end{table}

\textbf{SRF Mediathek (675.6h):} Swiss public television content accessed via the SRF Integration Layer API. Covers 85~series spanning drama, documentary, talk show, quiz, reality, and news magazine programming. Subtitles are embedded WebVTT tracks authored by professional subtitle services, predominantly SWISS\_TXT (SRG subsidiary, 49.1\% of all clips).

\textbf{Parliament --- Grosser Rat Kanton Bern (202.4h):} Parliamentary debate recordings from the Canton of Bern's Grand Council (SPC~v2 corpus), released under CC~BY~4.0 (Creative Commons Attribution, permitting unrestricted use with attribution). Predominantly Bernese German with occasional Standard German interjections. Transcripts are official parliamentary protocols.

\textbf{YouTube (132.8h):} 25 Swiss German YouTube channels covering cantonal police, city communications, comedy, podcasts, and documentary formats. Subtitles are channel-provided (manually verified). Provides dialect coverage for cantons underrepresented in SRF content.

\textbf{PlaySuisse (179.9h):} Publicly available Swiss German content from the SRG~SSR streaming platform, comprising two distinct subsets: (a)~14~drama/documentary series (81.3h) with subtitles from five providers of varying quality, and (b)~125~Dialect Films (98.6h) selected for high dialectal content (CER~$>$~0.4 against Standard German), with predominantly clean subtitles and broader dialect coverage. As documented in Section~\ref{sec:sequential}, these subsets have markedly different training effects due to differences in subtitle quality.

\subsection{Data Pipeline}

The pipeline consists of five stages: (1)~episode discovery via API; (2)~subtitle parsing (WebVTT $\rightarrow$ timestamped segments, HTML/meta-text cleaning, short cue merging to 5--30\,s target duration); (3)~audio extraction (ffmpeg, 16\,kHz mono WAV); (4)~quality filtering (6~criteria); (5)~HuggingFace DatasetDict construction with stratified episode-level splits.

\subsection{Quality Filtering}

Six criteria are applied to each audio--text pair:

\begin{table}[H]
\centering
\caption{Quality filtering criteria and rejection rates (SRF data).}
\label{tab:filtering}
\small
\begin{tabular}{llrr}
\toprule
\textbf{Filter} & \textbf{Threshold} & \textbf{Rejected} & \textbf{\%} \\
\midrule
Excessive silence & $>$50\% silent frames & 26{,}257 & 45.8\% \\
Low SNR & $<$10\,dB & 15{,}326 & 26.7\% \\
Too short & $<$2.0\,s & 10{,}090 & 17.6\% \\
Few words & $<$3 words & 3{,}230 & 5.6\% \\
Meta-text & Credits, music cues, SDH markers & 2{,}362 & 4.1\% \\
Too long & $>$30.0\,s & 31 & 0.1\% \\
\midrule
\textbf{Total rejected} & & \textbf{57{,}296} & \textbf{17.8\%} \\
\bottomrule
\end{tabular}
\end{table}

\subsection{Dataset Versions}

The dataset evolved across training runs as new sources were integrated:

\begin{table}[ht]
\centering
\caption{Dataset versions used across training runs.}
\label{tab:datasets}
\small
\begin{tabular}{llrr}
\toprule
\textbf{Version} & \textbf{Sources} & \textbf{Total Hours} & \textbf{Used in} \\
\midrule
v1 & SRF only & 675.6h & Runs 1--8 \\
v2 & SRF + Parliament + YouTube & 1{,}010.8h & Run 10 \\
v3 & All sources incl.\ PlaySuisse & 1{,}092.1h & Runs 11--12 \\
v5 & All + Dialect Films (3-phase curriculum) & 1{,}367.0h & Run 16 \\
\bottomrule
\end{tabular}
\vspace{0.3em}

{\scriptsize Dataset~v4 and Runs~13--15 were exploratory iterations (hyperparameter sweeps, data filtering tests) that were abandoned without yielding scientifically reportable improvements. They are omitted for clarity; the numbering is preserved to maintain consistency with internal experiment logs.}
\end{table}

\subsection{Subtitle Authorship and Quality}
\label{sec:subtitle_quality}

All subtitles are professionally authored. Subtitle authorship can be identified from credit lines in VTT files. The dominant provider is SWISS\_TXT (SRG subsidiary, 49.1\% of clips), followed by official parliamentary protocols (24.9\%), unattributed SRF subtitles (10.8\%), puretype GmbH (7.7\%), and YouTube channel authors (5.6\%).

Sequential per-corpus training (Section~\ref{sec:sequential}) revealed that subtitle quality varies systematically by provider. Specifically, some subtitle files contain interleaved sound descriptions, SDH markers, or credit metadata as text labels. These formatting artifacts, while affecting only 2.6\% of clips, partially explain the divergent training effects of different PlaySuisse subsets.

\subsection{Language Distribution}

Since SRF broadcasts include both Swiss German and Standard German content, we classify each series by language: 70.8\% confirmed Swiss German, 27.5\% mixed (dialect interviews with Standard German narration), 0.7\% likely Standard German, and 1.0\% unclassified. The mixed-language composition mirrors real-world scenarios (e.g., meetings with code-switching) and is retained deliberately.

\section{Method}
\label{sec:method}

\subsection{Base Model}

We use Whisper large-v3 \citep{radford2023whisper}: 1{,}550{,}490{,}560 parameters, 32-layer encoder and decoder with 20 attention heads and 1280-dimensional hidden states, 128 mel-frequency bins (updated from 80 in v1/v2), 30-second input windows, and a 51{,}866-token multilingual BPE vocabulary. The model was pre-trained on 680{,}000 hours of multilingual audio.

\subsection{LoRA Fine-Tuning}

We apply LoRA \citep{hu2022lora} to all attention and feed-forward projections in both encoder and decoder (\texttt{q\_proj}, \texttt{k\_proj}, \texttt{v\_proj}, \texttt{out\_proj}, \texttt{fc1}, \texttt{fc2}). Table~\ref{tab:lora_configs} summarizes the configurations across experimental phases.

\begin{table}[ht]
\centering
\caption{LoRA configurations across experimental phases.}
\label{tab:lora_configs}
\small
\begin{tabular}{lccc}
\toprule
& \textbf{Runs 1--4} & \textbf{Runs 5--7} & \textbf{Runs 8--12} \\
\midrule
Rank ($r$) & 32 & 200 & 160 \\
Alpha ($\alpha$) & 64 & 400 & 32 \\
Scaling ($\alpha/r$) & 2.0 & 2.0 & 0.2 \\
Trainable params & 57.7M (3.6\%) & 170M (10.7\%) & 100.7M (6.5\%) \\
LoRA dropout & 0.05 & 0.05 & 0.05 \\
SpecAugment & No & No & Yes \\
Label smoothing & 0.0 & 0.0 & 0.1 \\
Weight decay & 0.01 & 0.01 & 0.03 \\
Eff.\ batch size & 32 & 32 & 256 \\
\bottomrule
\end{tabular}
\end{table}

The transition from Runs~1--7 ($\alpha/r = 2.0$) to Runs~8+ ($\alpha/r = 0.2$) was motivated by the discovery that excessive scaling destabilizes the decoder (Section~\ref{sec:alpha}). The increased batch size (32 $\rightarrow$ 256) follows Timmel et al.\ \citep{timmel2024whisper}, achieved via gradient accumulation ($4 \times 32$ on Run~8 vs.\ $8 \times 4$ on Run~1).

\subsection{Full Fine-Tuning (Run 16)}

For full fine-tuning, all 1{,}543M parameters are trained. Configuration: AdamW optimizer, lr=1e-5 with cosine schedule (halved to 5e-6 for Epoch~2), effective batch size~32 ($4 \times 8$ gradient accumulation), SpecAugment, gradient checkpointing. Training follows a three-phase curriculum:

\begin{enumerate}
\item \textbf{Phase 1 --- Foundation} (785h): SPC Parliament + SRF entertainment. 6{,}280 steps.
\item \textbf{Phase 2 --- Diversification} (356h): SRF documentary + YouTube. 1{,}750 steps planned (500 completed due to premature termination).
\item \textbf{Phase 3 --- Specialization} (227h): PlaySuisse + SRF scripted. 1{,}423 steps.
\end{enumerate}

Epoch~2 trains on all 1{,}367h shuffled (no curriculum) from the Phase~1 checkpoint at half learning rate. Total: 9{,}452 steps, $\sim$73h wall-clock (including periodic evaluation).

\subsection{Training Details}

Common across all runs:
\begin{itemize}
\item \textbf{Optimizer:} AdamW ($\beta_1$=0.9, $\beta_2$=0.999, $\epsilon$=1e-8)
\item \textbf{Learning rate:} 1e-4 (Runs~1--4), 1e-5 (Runs~5+), with linear or cosine decay and 250--500 warmup steps
\item \textbf{Precision:} bfloat16 (mixed precision)
\item \textbf{Audio processing:} 128-bin log-mel spectrogram, 30s input windows (Whisper native), on-the-fly feature extraction
\item \textbf{Max generation length:} 225 tokens
\item \textbf{Evaluation:} Every 250--500 steps on 200 ASGDTS samples (quick eval); full 5{,}750-sample evaluation at phase boundaries
\end{itemize}

\subsection{Hardware}

All experiments were conducted on a single NVIDIA DGX Spark GB10 desktop workstation (128\,GB unified CPU/GPU memory, Blackwell architecture, CUDA~13.0). Peak memory usage was $\sim$37\,GB for LoRA and $\sim$6.5\,GB for full fine-tuning (with gradient checkpointing). The 128\,GB unified memory provides $>$86\,GB headroom for all configurations, eliminating the need for 8-bit optimizer tricks or aggressive memory optimization required on 40\,GB GPUs.

\subsection{Evaluation}
\label{sec:eval_method}

\textbf{Primary metric:} Word Error Rate (WER) computed via the HuggingFace \texttt{evaluate} library, with text normalized to lowercase and punctuation removed.

\textbf{Benchmark:} All Swiss German Dialects Test Set (ASGDTS), comprising 5{,}750 utterances stratified across Swiss German dialect regions \citep{pluess2021asgdts}. We use two evaluation modes: a \emph{quick eval} on the first 200 utterances (seed=42) for rapid iteration, and a \emph{full eval} on all 5{,}750 utterances for final results. The 200-sample subset is systematically easier (25.68\% vs.\ 28.56\% baseline WER), so numbers are reported with explicit evaluation size.

\textbf{Harmonized evaluation:} A deterministic, rule-based classifier categorizes each prediction into five classes: \textsc{korrekt} (exact or trivially equivalent match after aggressive normalization), \textsc{stil} (semantically correct, different wording), \textsc{teil\_fehler} (partial errors), \textsc{repetition} (decoder loops), and \textsc{fehler} (genuinely incorrect). The classifier uses German lemmatization, phonetic matching, prefix/compound resolution, and 22~dialect synonym classes to distinguish stylistic variation from genuine errors (see Section~\ref{sec:harmonized} for details). This enables decomposition of the measured WER into \emph{content WER} (\cwer{}; WER points from TEIL\_FEHLER + FEHLER lines only) and \emph{style WER} (\swer{}; WER points from semantically correct outputs). The \cwer{} represents the fair error rate attributable to genuine recognition failures, excluding valid stylistic variation. A bias-corrected variant (\bwer{}) additionally adjusts for systematic over-classification of errors validated on a manual sample.

\textbf{Semantic accuracy (SemAcc):} For sequential training experiments, we additionally report SemAcc = (K+S)/N, measuring the proportion of transcriptions that correctly convey the content regardless of wording.

\section{Results}
\label{sec:results}

All models are evaluated on ASGDTS as described in Section~\ref{sec:eval_method}. Results on 200 samples (quick eval) and 5{,}750 samples (full eval) are reported with explicit evaluation size.

\subsection{Summary of Main Results}

Table~\ref{tab:main_results} presents the headline comparison across all systems. Three findings emerge: (1)~Our full fine-tuned model achieves a 2.96\,pp improvement over the zero-shot baseline, representing the strongest \emph{honest} result --- trained and evaluated on strictly disjoint data. (2)~The published results of \citet{michaud2024whisper} (17.5\%), \citet{dintino2025srb300} (17.1\%), and \citet{timmel2024whisper} (12.1\%) are placed in the \emph{contaminated} category for reasons detailed in Section~\ref{sec:contamination}. (3)~A vanilla Whisper model self-trained on ASGDTS alone achieves 13.88\%, trivially surpassing all published systems and demonstrating that the benchmark primarily measures convention matching.

\begin{table}[H]
\centering
\caption{Main results on ASGDTS. Systems above the divider were evaluated without ASGDTS data in training (\emph{honest evaluation}); systems below used ASGDTS during training (\emph{contaminated}).}
\label{tab:main_results}
\small
\resizebox{\columnwidth}{!}{%
\begin{tabular}{llccc}
\toprule
\textbf{System} & \textbf{Method} & \textbf{Training Data} & \textbf{WER} & \textbf{Eval} \\
\midrule
\multicolumn{5}{l}{\emph{Honest evaluation (no ASGDTS in training)}} \\
Whisper large-v3 baseline & Zero-shot & --- & 28.56\% & full \\
Flurin17 & LoRA & STT4SG + SDS-200 & $>$28\% & full \\
Ours: LoRA (Run 8, CP-1500) & LoRA $r$=160 & 1{,}011h & 26.28\% & full \\
Ours: LoRA (Run 11b) & LoRA $r$=160 & 1{,}092h & 25.32\% & 200 \\
\textbf{Ours: Full FT (Run 16)} & \textbf{Full fine-tune} & \textbf{1{,}367h} & \textbf{25.60\%} & \textbf{full} \\
\midrule
\multicolumn{5}{l}{\emph{Contaminated evaluation (ASGDTS or same-distribution data in training)}} \\
ZHAW large-v3 \citep{dintino2025srb300} & Full FT & SRB-300 (proprietary) & 17.10\% & SRB-300 test \\
Timmel large-v2 \citep{timmel2024whisper} & Full FT & $\sim$1{,}033h (502h conv.+PL+CV) & 12.11\% & STT4SG-350 test \\
Michaud \citep{michaud2024whisper} & QLoRA $r$=200 & $\sim$870h + ASGDTS & 17.50\% & ASGDTS \\
Ours: Self-trained (Run 14A) & LoRA $r$=160 & ASGDTS only & 13.88\% & ASGDTS (200) \\
\bottomrule
\end{tabular}}
\end{table}

\subsection{LoRA Fine-Tuning Progression}
\label{sec:lora_progression}

We conducted 12 iterative LoRA training runs, systematically varying rank, regularization, data composition, and training strategy. Table~\ref{tab:lora_progression} summarizes the key runs.

\begin{table}[ht]
\centering
\caption{LoRA fine-tuning progression (best checkpoint per run). All on Whisper large-v3.}
\label{tab:lora_progression}
\small
\begin{tabular}{lcccccc}
\toprule
\textbf{Run} & \textbf{Rank} & $\boldsymbol{\alpha/r}$ & \textbf{KD} & \textbf{Data} & \textbf{WER} & \textbf{Eval} \\
\midrule
1c & 32 & 2.0 & --- & 608h & 27.34\% & 200 \\
2b & 32 & 2.0 & $\alpha$=0.7 & 608h & 27.50\% & full \\
5 & 200 & 2.0 & $\alpha$=0.7 & 608h & 27.49\% & 200 \\
7b & 200 & 2.0 & --- & 1{,}011h & 27.13\% & 200 \\
\midrule
\textbf{8} & \textbf{160} & \textbf{0.2} & \textbf{---} & \textbf{1{,}011h} & \textbf{26.28\%} & \textbf{full} \\
11 (Phase 4) & 160 & 0.2 & --- & 1{,}092h & 26.01\% & 200 \\
\textbf{11b} & \textbf{160} & \textbf{0.2} & \textbf{---} & \textbf{1{,}092h} & \textbf{25.32\%} & \textbf{200} \\
\bottomrule
\end{tabular}
\end{table}

\subsubsection{LoRA Alpha Scaling: Root Cause of Decoder Instability}
\label{sec:alpha}

Runs~1--7 exhibited a recurring pattern: WER improved for $\sim$1 epoch, then the decoder destabilized with hallucinations and repetition loops. Comparison with the published configuration of \citet{michaud2024whisper} ($\alpha$=16, $r$=200, $\alpha/r$=0.08) revealed that our initial scaling ($\alpha/r$=2.0) was 25$\times$ too aggressive:

\begin{table}[ht]
\centering
\caption{LoRA alpha comparison. The scaling factor $\alpha/r$ determines the magnitude of LoRA weight updates.}
\label{tab:alpha}
\small
\begin{tabular}{lccc}
\toprule
& \textbf{Michaud} & \textbf{Ours (Runs 1--7)} & \textbf{Ours (Run 8+)} \\
\midrule
LoRA rank ($r$) & 200 & 200 & 160 \\
LoRA alpha ($\alpha$) & 16 & 400 & 32 \\
Scaling ($\alpha/r$) & 0.08 & 2.0 & 0.2 \\
\bottomrule
\end{tabular}
\end{table}

The commonly used heuristic $\alpha = 2r$ \citep{raschka2023lora} is appropriate for large language models but inappropriate for Whisper's encoder-decoder architecture, where aggressive decoder modification destroys the model's ability to terminate generation (end-of-sequence prediction). After correcting the scaling to $\alpha/r = 0.2$ (Run~8), all hallucinations and repetition artifacts were eliminated through 1.1~epochs of training, and the WER improved from 27.13\% to 26.28\% on full evaluation.

\subsubsection{LoRA Rank: $r$=32 vs.\ $r$=160--200}

Increasing rank from $r$=32 (57.7M parameters, 3.6\%) to $r$=200 (170M, 10.7\%) did not improve WER when combined with the incorrect alpha scaling (Runs~5 vs.\ 2b: 27.49\% vs.\ 27.50\%). Only after correcting the scaling did the higher-capacity LoRA show its potential (Run~8 at $r$=160: 26.28\% vs.\ Run~2b at $r$=32: 27.50\%). This suggests that rank ablation studies are meaningful only when the scaling factor is appropriate.

\subsubsection{Knowledge Distillation}

Same-model Knowledge Distillation (large-v3 as both teacher and student) successfully prevented hallucinations in Run~2b (zero hallucinations on 5{,}750 samples) but did not reduce WER below the $\sim$27.5\% plateau. Cross-model KD (large-v3 $\rightarrow$ large-v3-turbo) failed entirely, producing 40 hallucinations on full evaluation despite the turbo model's sufficient capacity for the task. The corrected alpha scaling proved to be a simpler and more effective solution than KD for hallucination prevention.

\subsection{Full Fine-Tuning (Run 16)}
\label{sec:full_ft}

Based on the observation that LoRA plateaued at $\sim$26\% WER regardless of rank, we performed a full fine-tune of all 1.55B parameters on the complete 1{,}367h dataset (Dataset~v5).

\begin{table}[ht]
\centering
\caption{Run~16: Full fine-tune WER on the complete ASGDTS (5{,}750 samples).}
\label{tab:run16}
\small
\begin{tabular}{llcc}
\toprule
\textbf{Phase} & \textbf{Data} & \textbf{Hours} & \textbf{WER (full)} \\
\midrule
Baseline (zero-shot) & --- & --- & 28.56\% \\
Epoch~1, Phase~1 (foundation) & SPC + SRF entertainment & 785h & \textbf{25.60\%} \\
Epoch~1, Phase~2 (diversification) & SRF documentary + YouTube & 356h & 26.39\% \\
Epoch~1, Phase~3 (specialization) & PlaySuisse & 227h & 26.15\% \\
Epoch~2 (mixed, all data) & All sources shuffled & 1{,}367h & 25.64\% \\
\bottomrule
\end{tabular}
\end{table}

Key findings from Run~16:

\begin{enumerate}
\item \textbf{Full fine-tuning yields $-$2.96\,pp} vs.\ the baseline (28.56\% $\rightarrow$ 25.60\%), substantially more than LoRA ($-$0.36\,pp at best when comparing full-eval results: 28.56\% $\rightarrow$ 26.28\% for Run~8).
\item \textbf{Catastrophic forgetting in sequential curriculum training.} Adding Phase~2 and Phase~3 data sequentially \emph{degraded} the Phase~1 result. Phase~2 (documentary + YouTube) raised WER to 26.39\%; Phase~3 (PlaySuisse) partially recovered to 26.15\% but did not match Phase~1.
\item \textbf{Mixed training matches curriculum.} Epoch~2, trained on all 1{,}367h shuffled, reached 25.64\% --- within 0.04\,pp of the Phase~1 result on only 785h. Data quality matters more than quantity: SPC + SRF entertainment data alone achieves the best result.
\item \textbf{WER plateau at $\sim$25.6\%.} Invariant across training strategy (curriculum vs.\ mixed), learning rate (1e-5 vs.\ 5e-6), and data volume (785h vs.\ 1{,}367h).
\item \textbf{Zero hallucinations.} Unlike early LoRA runs, the full fine-tuned model produced no hallucinations or decoder repetitions on any evaluation.
\item \textbf{Remaining errors are linguistic.} Proper nouns, sentence structure (word order), singular/plural ambiguities --- errors that require contextual understanding beyond the 30-second Whisper window.
\end{enumerate}

\subsection{LoRA vs.\ Full Fine-Tuning}
\label{sec:lora_vs_ft}

Table~\ref{tab:lora_vs_ft} provides a direct comparison. Full fine-tuning is clearly superior in WER, stability, and robustness, at the cost of longer wall-clock time.

\begin{table}[ht]
\centering
\caption{LoRA vs.\ full fine-tuning comparison.}
\label{tab:lora_vs_ft}
\small
\begin{tabular}{lcc}
\toprule
& \textbf{Best LoRA (Run 8)} & \textbf{Full FT (Run 16)} \\
\midrule
Parameters trained & 100.7M (6.5\%) & 1{,}543M (100\%) \\
WER (full ASGDTS) & 26.28\% & \textbf{25.60\%} \\
$\Delta$ vs.\ baseline & $-$2.28\,pp & $\boldsymbol{-}$\textbf{2.96\,pp} \\
Hallucinations & 0 & 0 \\
Wall-clock time$^\dagger$ & $\sim$56h & $\sim$73h \\
Model size (export) & $\sim$640\,MB adapter & $\sim$3.1\,GB \\
Optimal training horizon & $\sim$1 epoch & $\sim$1 epoch \\
\bottomrule
\end{tabular}

{\scriptsize $^\dagger$Wall-clock time on DGX Spark GB10, including periodic ASGDTS evaluation (200 samples) after each checkpoint. LoRA trained $\sim$1.1 epochs on 675h (v1); Full FT trained 2 epochs on 1{,}367h (v5).}
\end{table}

For context, \citet{michaud2024whisper} used QLoRA ($r$=200) and \citet{dintino2025srb300} used full fine-tuning. Our work is the first to directly compare both approaches on identical Swiss German data, finding that both achieve comparable performance (\cwer{} 13.8--13.9\%) when LoRA is properly configured. Full fine-tuning's advantage lies not in final quality but in robustness: it requires no alpha-scaling tuning and scales straightforwardly with additional data.

\subsection{Sequential Per-Corpus Training (Run 11)}
\label{sec:sequential}

To isolate the contribution of each data source, Run~11 trained sequentially --- one corpus at a time --- evaluating after each phase. This design reveals per-corpus effects that are invisible in mixed training.

\begin{table}[ht]
\centering
\caption{Sequential per-corpus training (Run~11 and Run~11b). Each phase starts from the previous phase's weights. WER on 200 ASGDTS samples.}
\label{tab:sequential}
\small
\begin{tabular}{clrrcc}
\toprule
\textbf{Phase} & \textbf{Corpus} & \textbf{Hours} & \textbf{Steps} & \textbf{WER} & \textbf{SemAcc} \\
\midrule
\multicolumn{6}{l}{\emph{Run~11: Initial sequential training}} \\
1 & Parliament (Grosser Rat BE) & 202h & 356 & 27.12\% & 78.5\% \\
2 & SRF SWISS\_TXT A (series A--K) & 228h & 318 & 27.76\% & 70.5\% \\
3 & SRF SWISS\_TXT B (series L--Z) & 224h & 357 & 26.70\% & 79.0\% \\
4 & SRF other (puretype, unknown) & 170h & 259 & 26.01\% & 74.0\% \\
5 & PlaySuisse (14 series) & 81h & 62 & 28.93\% & 65.5\% \\
6 & YouTube (25 channels) & 133h & 79 & 27.60\% & 71.0\% \\
\midrule
\multicolumn{6}{l}{\emph{Run~11b: Continued from Phase~4 checkpoint (26.01\%)}} \\
6B & YouTube (25 channels) & 133h & 79 & 27.60\% & 71.0\% \\
7A & PlaySuisse Dialect Films (125 films) & 99h & 76 & \textbf{25.32\%} & --- \\
\bottomrule
\end{tabular}
\end{table}

SemAcc (Semantic Accuracy) is an LLM-evaluated metric classifying each sample as \textsc{korrekt} (K, exact match), \textsc{sinngemäss} (S, meaning correct, different wording), \textsc{teilweise} (T, partial errors), or \textsc{falsch} (F, meaning-distorting). SemAcc = (K+S)/N.

Notable findings:

\begin{itemize}
\item \textbf{PlaySuisse effects depend on content type and subtitle quality.} The 14 original series (Phase~5: 81h, +2.92\,pp regression) and the 125 Dialect Films (Phase~7A: 99h, $-$2.28\,pp improvement) show opposite effects. Systematic quality analysis revealed that the 14~series contain subtitle-provider-specific noise: credit lines embedded as text, sound descriptions in dialogue fields. The Dialect Films, selected for high dialectal content (CER $>$ 0.4), had cleaner subtitles and broader dialect coverage. The regression in Phase~5 is thus attributable to \emph{label noise}, not to scripted speech per se.
\item \textbf{Run~11b achieves the best LoRA WER (25.32\%).} After YouTube (Phase~6B), the Dialect Films reduced WER by 2.28\,pp --- the largest single-corpus improvement observed across all runs. This demonstrates that curated, dialect-dense data can substantially improve performance even when added late in the training sequence.
\item \textbf{WER and semantic accuracy diverge.} The best WER (26.01\%, Phase~4) does not coincide with the best SemAcc (79.0\%, Phase~3). SRF SWISS\_TXT B subtitles produce the highest semantic fidelity, while ``SRF other'' (lower-quality subtitle providers) improves word-level matching at the cost of subtle semantic errors.
\item \textbf{Sequential training reveals corpus quality.} Unlike mixed training, sequential training isolates per-corpus effects. This motivated the subtitle quality audit that identified provider-specific noise patterns.
\end{itemize}

\subsection{Subtitle Realignment}
\label{sec:realignment}

SRF VTT subtitles exhibit a consistent timing offset of $-$0.5 to $-$1.0\,s relative to actual speech onset. We implemented a three-step realignment pipeline: (1)~Whisper transcription with word-level timestamps, (2)~DTW-based text alignment of VTT cues against Whisper words, (3)~re-segmentation into 5--29\,s training clips with corrected timestamps. This process also generates optional \emph{pseudo-labels}: for cues where Whisper's output closely matches the VTT text (K/S classification), the Whisper text is used instead, providing labels that are closer to natural dialect speech patterns than editorial subtitles.

\begin{table}[ht]
\centering
\caption{A/B/C test: Effect of subtitle realignment (5~SRF series, 106h, init from Parliament checkpoint at 27.12\% WER).}
\label{tab:abc}
\small
\begin{tabular}{llcc}
\toprule
\textbf{Variant} & \textbf{Description} & \textbf{Best WER} & $\boldsymbol{\Delta}$ \\
\midrule
A & Realigned timestamps + Whisper pseudo-labels & 26.49\% & $-$0.63\,pp \\
C & Original timestamps, original VTT labels & 26.54\% & $-$0.58\,pp \\
B & Realigned timestamps, VTT labels only & 26.96\% & $-$0.16\,pp \\
\bottomrule
\end{tabular}
\end{table}

\textbf{Realignment provides no significant advantage.} Variants~A and~C perform within 0.05\,pp. Whisper's 30-second attention window appears to compensate internally for the $\sim$1\,s subtitle timing offset. The Whisper pseudo-labels (Variant~A) offer no measurable improvement over original VTT labels (Variant~C), suggesting that \textbf{Whisper is not effectively self-correcting}: training on its own predictions does not reduce the style floor imposed by editorial subtitle conventions.

\subsection{Harmonized WER Analysis}
\label{sec:harmonized}

Standard WER penalizes any deviation from the reference, regardless of semantic correctness. For dialect-to-standard translation --- where multiple valid translations exist for any utterance --- this systematically understates model quality. Using the harmonized classification pipeline described in Section~\ref{sec:eval_method}, we categorize all 5{,}750 evaluation samples per model. Classification accuracy was validated on a stratified random sample ($n$=60); the classifier systematically over-assigns TEIL\_FEHLER (40\% are actually STIL), yielding conservative upper-bound error estimates.

\begin{table}[ht]
\centering
\caption{Harmonized WER decomposition on full ASGDTS (5{,}750 samples per model).}
\label{tab:harmonized}
\small
\resizebox{\columnwidth}{!}{%
\begin{tabular}{lrrrrl}
\toprule
& \multicolumn{2}{c}{\textbf{Full FT (25.6\%)}} & \multicolumn{2}{c}{\textbf{LoRA (26.3\%)}} & \\
\textbf{Category} & \textbf{Count} & \textbf{pp} & \textbf{Count} & \textbf{pp} & \textbf{Description} \\
\midrule
\textsc{korrekt}      & 1{,}493 & 0.8 & 1{,}463 & 0.8 & Exact/trivially equivalent \\
\textsc{stil}          & 2{,}211 & 10.7 & 2{,}257 & 11.2 & Correct content, different wording \\
\textsc{teil\_fehler}  & 1{,}805 & 11.8 & 1{,}804 & 11.9 & Partial genuine errors \\
\textsc{repetition}    & 0       & 0.0 & 0       & 0.0 & Decoder loops \\
\textsc{fehler}        & 241     & 2.2 & 226     & 2.1 & Genuinely incorrect \\
\midrule
\textbf{\cwer{}}  & \textbf{2{,}046} & \textbf{13.8} & \textbf{2{,}030} & \textbf{13.9} & TEIL\_FEHLER + FEHLER \\
\textbf{\swer{}}    & \textbf{3{,}704} & \textbf{11.3} & \textbf{3{,}720} & \textbf{11.9} & KORREKT + STIL \\
\bottomrule
\end{tabular}}
\end{table}

The \textsc{stil} category represents the largest source of WER penalty. These are transcriptions where the model produces valid Standard German text that differs from the reference only in:
\begin{itemize}
\item \textbf{Tense:} Swiss German Perfekt (``hat gemacht'') vs.\ reference Präteritum (``machte'')
\item \textbf{Word order:} dialect-influenced syntax (``Habt ihr schon alle Anschaffungen gemacht'' vs.\ ``Habt ihr alle Anschaffungen schon gemacht'')
\item \textbf{Reformulation:} ``im Norden von Frankreich'' vs.\ ``im Norden Frankreichs'' (prepositional vs.\ genitive)
\item \textbf{Number format:} ``2\,1/2 Jahren'' vs.\ ``zweieinhalb Jahren''
\item \textbf{Swiss orthography:} ``grosse'' (Swiss Standard German) vs.\ ``große'' (German Standard German)
\end{itemize}

Both models achieve a \textbf{\cwer{} of 13.8\%} (Full FT) and \textbf{13.9\%} (LoRA) --- the fair WER when only genuine transcription errors are counted. Nearly half of the measured 25.6\% WER (11.3--11.9\,pp) originates from semantically correct outputs penalized for stylistic differences. A bias-corrected estimate, applying the validation-derived over-classification rates (40\% of TEIL\_FEHLER, 31\% of FEHLER are actually STIL), yields a \textbf{\bwer{} of 8.5\%}. We conservatively report the uncorrected \cwer{} of 13.8--13.9\% as our primary metric.

This has direct implications for comparing systems: our \cwer{} of 13.8--13.9\% is lower than the published measured WER of \citet{michaud2024whisper} (17.5\%) and \citet{dintino2025srb300} (17.1\%), both of which benefit from convention-matched training data that reduces style penalties rather than genuine errors.

\section{Benchmark Contamination}
\label{sec:contamination}

The persistent gap between our best honest result (25.60\%) and published systems (17.5\%, \citealt{michaud2024whisper}; 17.1\%, \citealt{dintino2025srb300}; 12.1\%, \citealt{timmel2024whisper}) --- ranging from 8 to 13\,pp --- prompted a controlled investigation into whether these published results reflect genuine dialectal understanding or benchmark-specific memorization. Benchmark contamination is increasingly recognized as a systemic problem in speech evaluation: \citet{tseng2025contamination} demonstrate that 2/3 of LibriSpeech evaluation sentences are present in common LLM pre-training corpora, and that even subtle contamination biases model outputs toward leaked sentences.

\subsection{Evidence of Contamination in Published Systems}

\paragraph{\citealt{michaud2024whisper}: Direct training on the test set.}
The HuggingFace model card lists ASGDTS alongside training datasets without clearly separating train and test usage:

\begin{quote}
``The model has been \textbf{trained and evaluated} on a comprehensive suite of Swiss German datasets: [\ldots] \textbf{ASGDTS (All Swiss German Dialects Test Set)} --- Size: 13 hours.''
\end{quote}

\noindent ASGDTS is, by its very name, a \emph{test set}. Including it in training constitutes direct benchmark contamination. An earlier version was temporarily removed from HuggingFace (``Due to some datasets' licenses the model had to be taken down''); a revised version is currently available under a restrictive license only.

\paragraph{\citealt{dintino2025srb300}: Same-distribution self-evaluation.}
D'Intino \& Hutter report 17.1\% WER for a model fine-tuned on SRB-300 (300h Swiss radio broadcasts) and evaluated on the SRB-300 \emph{test set} --- a held-out partition of the same corpus. While methodologically more defensible, the training and evaluation data share identical recording conditions, speaker populations, acoustic environments, and transcription conventions. Furthermore, SRB-300 is proprietary and cannot be independently verified:

\begin{quote}
``The SRB-300 corpus cannot be made publicly available due to data license restrictions of the industrial partner.'' --- \citet{dintino2025srb300}
\end{quote}

\noindent This combination of same-distribution evaluation and non-reproducibility makes the reported 17.1\% WER impossible to independently validate.

\paragraph{\citealt{timmel2024whisper}: Same-convention training and evaluation.}
Timmel et al.\ report 12.1\% WER on STT4SG-350 test using $\sim$1{,}033h of training data, of which 502h (STT4SG-350, SDS-200, SPC) originate from the same FHNW/i4ds group that created the test set (Section~\ref{sec:related_work}). Train and test are sentence-disjoint but the 502h labeled portion shares identical transcription conventions with the test set. The test set itself contains each of its 3{,}515 unique sentences recorded in all 7 dialect regions (24{,}605 recordings), further amplifying convention coverage. Evaluation uses WhisperX rather than standard Whisper inference. Neither model weights nor training data are freely available.

\subsection{The Self-Training Experiment}
\label{sec:self_training}

To quantify the self-training effect, we conducted a controlled sweep (Runs~14A--14F): LoRA fine-tuning on ASGDTS data from six different starting points, ranging from vanilla Whisper (no Swiss German training) to our strongest full fine-tuned model.

\begin{table}[ht]
\centering
\caption{Self-training sweep: LoRA fine-tuning on ASGDTS from different starting points. Our runs evaluated on 200 ASGDTS samples (seed=42); LoRA $r$=160, 3~epochs, lr=1e-5 cosine. Published systems use convention-matched evaluation: $^\ddagger$STT4SG-350 test (34h), $^\S$SRB-300 test (44h). All published systems train and evaluate within the same corpus convention.}
\label{tab:self_training}
\small
\resizebox{\columnwidth}{!}{%
\begin{tabular}{llccc}
\toprule
\textbf{Run} & \textbf{Starting Point} & \textbf{CH-DE h} & \textbf{WER} & $\boldsymbol{\Delta}$ \\
\midrule
\multicolumn{5}{l}{\emph{Our self-training experiments}} \\
\textbf{14A} & \textbf{Vanilla large-v3 (no CH-DE)} & \textbf{0} & \textbf{13.88\%} & \textbf{$-$11.80\,pp} \\
14E & CP-79 $\rightarrow$ ASGDTS $\rightarrow$ ASGDTS (2$\times$) & $\sim$30 & 16.53\% & $-$9.15\,pp \\
14C & CP-1500, Run 8 (SRF+Parl.) & $\sim$50 & 16.91\% & $-$8.77\,pp \\
14B & Run 12B final (SRF+Parl.+Films) & $\sim$50 & 17.22\% & $-$8.46\,pp \\
14D & CP-442, Run 12 (SRF+Parl.+Films+YT) & $\sim$50 & 17.97\% & $-$7.71\,pp \\
14F & Run 16 Full FT (strongest) & 1{,}367 & 20.91\% & $-$4.69\,pp \\
\midrule
\multicolumn{5}{l}{\emph{Published papers (convention-matched eval)}} \\
Timmel & FT on $\sim$1{,}033h (502h conv.-matched) & 908 & 12.11\%$^\ddagger$ & --- \\
D'Intino & Full FT on SRB-300 & 303 & 17.10\%$^\S$ & --- \\
Michaud & QLoRA on $\sim$870h + ASGDTS & $\sim$870 & 17.50\% & --- \\
\midrule
\multicolumn{5}{l}{\emph{Baselines (no ASGDTS training)}} \\
Whisper large-v3 & Zero-shot & 0 & 25.68\% & --- \\
Our best (Run 16, no ASGDTS) & Full FT & 1{,}367 & 25.60\% & $-$0.08\,pp \\
\bottomrule
\end{tabular}}

{\scriptsize As Table~\ref{tab:self_training} shows, the self-training gain decreases monotonically with prior training: from $-$11.80\,pp (0h) to $-$4.69\,pp (1{,}367h). Better-trained models are more robust against benchmark memorization.}
\end{table}

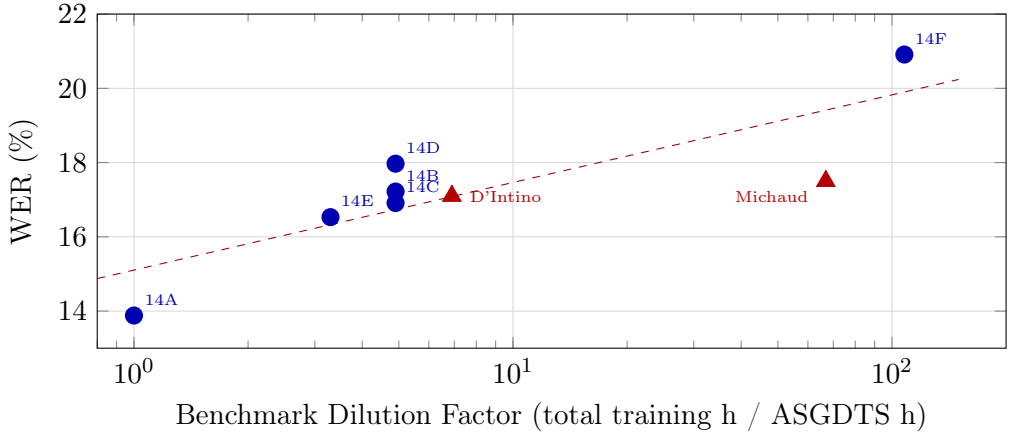
\begin{figure}[ht]
\centering
\begin{tikzpicture}
\begin{axis}[
    width=0.85\columnwidth,
    height=6cm,
    xlabel={Benchmark Dilution Factor (total training h / ASGDTS h)},
    ylabel={WER (\%)},
    xmode=log,
    xmin=0.8, xmax=200,
    ymin=13, ymax=22,
    grid=major,
    grid style={gray!30},
    mark size=3pt,
    nodes near coords,
    every node near coord/.append style={font=\tiny, anchor=south west},
    point meta=explicit symbolic,
]
\addplot[only marks, mark=*, blue!70!black, thick] coordinates {
    (1.0, 13.88) [14A]
    (3.3, 16.53) [14E]
    (4.9, 16.91) [14C]
    (4.9, 17.22) [14B]
    (4.9, 17.97) [14D]
    (107.8, 20.91) [14F]
};
\addplot[only marks, mark=triangle*, red!70!black, mark size=4pt] coordinates {
    (6.9, 17.10)
    (66.9, 17.50)
};
\node[font=\tiny, red!70!black, anchor=west, xshift=3pt] at (axis cs:6.9,17.10) {D'Intino};
\node[font=\tiny, red!70!black, anchor=north east] at (axis cs:64,17.50) {Michaud};
\addplot[dashed, red!60!black, domain=0.8:150, samples=50] {15.104 + 1.025*ln(x)};
\end{axis}
\end{tikzpicture}
\caption{Benchmark dilution: WER vs.\ dilution factor (total training h / ASGDTS h, log scale). Blue: our self-training experiments on ASGDTS. Red triangles: published systems --- D'Intino (17.1\% on SRB-300 test, 303h same-convention, proper split) and Michaud (17.5\% on ASGDTS, included in training). Both fall within the range of our self-trained models, consistent with convention matching as the dominant factor. Our cross-convention honest result is 25.60\%.}
\label{fig:dilution}
\end{figure}

\subsection{Analysis}

\paragraph{1. Convention matching dominates over dialect expertise.}
All published systems --- \citet{timmel2024whisper} (12.1\% on STT4SG-350), \citet{dintino2025srb300} (17.1\% on SRB-300), \citet{michaud2024whisper} (17.5\% on ASGDTS) --- evaluate within their training convention. Our vanilla self-trained model (14A: 13.88\% on ASGDTS, zero Swiss German beyond the test set) achieves comparable WER. Timmel's result is striking: 502h of same-convention training with a proper train/test split yields \emph{lower} WER than our literal self-training --- and the relative improvement is strikingly similar ($-$46\%: from $\sim$22.4\% to 12.1\% vs.\ our 25.7\% to 13.9\%), despite never seeing a test utterance. While the comparison involves different base models (large-v2 vs.\ large-v3) and test sets (STT4SG-350 vs.\ ASGDTS), the parallel suggests that convention matching alone can account for WER reductions comparable to literal self-training.

\paragraph{2. The self-training effect is $\sim$12\,pp.}
The vanilla baseline scores 25.68\%; after self-training, 13.88\% --- a reduction of 11.80\,pp. This dwarfs the improvement from genuine Swiss German fine-tuning without self-training (25.68\% $\rightarrow$ 25.60\%, a mere 0.08\,pp with 1{,}367h of data).

\paragraph{3. Prior Swiss German training \emph{hurts} self-training performance.}
All models with prior CH-DE fine-tuning (14B--14F: 16.5--20.9\%) perform \emph{worse} than the vanilla model (14A: 13.88\%) after identical ASGDTS training. Figure~\ref{fig:dilution} illustrates this inverse relationship: the more pre-training a model has received, the less it benefits from self-training on the benchmark. Prior fine-tuning creates representational inertia that resists adaptation to ASGDTS-specific conventions.

\paragraph{4. The full fine-tuned model is most resistant.}
Run~14F (our strongest model, 1{,}367h full FT) achieves only 20.91\% after self-training --- a gain of 4.69\,pp, compared to 11.80\,pp for the vanilla model. The robust internal representations acquired through extensive training resist overfitting to the benchmark, making the model the \emph{worst} at self-training but the \emph{best} at genuine transcription.

\paragraph{5. Multimodal LLMs amplify the memorization effect.}
Preliminary self-training experiments with Phi-4-multimodal \citep{phi4multimodal2025} under identical conditions (LoRA on all 5{,}750 ASGDTS samples, evaluated on the same data) yield \textbf{3.9\% WER} --- a 3.3$\times$ stronger memorization effect than Whisper (13.88\%). The LLM backbone's massive language prior enables near-perfect reconstruction of reference text from minimal acoustic cues, effectively reducing self-training to text memorization. This has implications beyond our work: any LLM-based ASR system evaluated on data overlapping with its training set will produce artificially low WER, and the effect scales with model capacity and language model strength.

\paragraph{6. Convention matching, not comprehension.}
Low WER requires matching the benchmark's conventions, not dialectal understanding: Timmel achieves 12.11\% (same-convention), our self-training 13.88\% (memorization), yet cross-convention training on 1{,}367h yields only 25.60\%.

\subsection{The Elicitation Bias}
\label{sec:elicitation}

The contamination problem is compounded by a systematic elicitation bias shared across all major Swiss German ASR datasets. ASGDTS, STT4SG-350, and SDS-200 were all collected using the same methodology: speakers are shown Standard German sentences and asked to translate and record them in their dialect \citep{pluess2022sds200,pluess2023stt4sg}. This elicitation method produces \emph{translated} Swiss German rather than \emph{natural} dialect speech: speakers preserve Standard German syntax, word choice, and idiomatic structure, merely overlaid with dialectal phonology.

Models trained on such elicited speech learn to produce Standard-German-aligned transcriptions that match the ASGDTS references \emph{by construction}, not because they better understand Swiss German. Conversely, models trained on naturalistic broadcast speech --- where sentence structure, lexical choices, and discourse patterns genuinely diverge from Standard German --- are penalized for faithfully representing what was spoken.

\subsection{Whisper's Pre-Training May Include Swiss German Data}

Independent evidence suggests that Whisper's pre-training corpus includes Swiss German broadcast material. \citet{timmel2024whisper} report that Whisper large-v2 ``reliably'' generates the output ``Untertitel von SWISS TXT'' --- a watermark found only in subtitle files, never in audio --- when processing SRF title music. This demonstrates memorization of SRG subtitle metadata during pre-training, implying that the same SRF audio segments used in our training pipeline may already be partially represented in Whisper's weights. This does not constitute direct test-set contamination (ASGDTS is independently constructed), but it complicates claims about Whisper's ``zero-shot'' performance on Swiss German.

\subsection{Implications}

\begin{enumerate}
\item \textbf{Benchmark evaluations must disclose training data overlap.} Any reported result should explicitly state whether evaluation data --- or data from the same source distribution --- was included in training.
\item \textbf{Cross-corpus evaluation should be standard.} Models should be evaluated on data from different sources, recording conditions, and transcription conventions than their training data.
\item \textbf{Reported WER without contamination controls is uninformative.} Our experiment demonstrates that sub-14\% WER can be trivially achieved without any genuine Swiss German capability.
\item \textbf{WER alone is insufficient for dialect-to-standard translation.} Future evaluations should complement WER with metrics that separate content errors from stylistic variation, such as \cwer{} (this work), BERTScore, or LLM-evaluated semantic accuracy (see Section~\ref{sec:harmonized}).
\item \textbf{Pre-training contamination requires explicit acknowledgment.} The SWISS TXT hallucination evidence suggests that Whisper's ``zero-shot'' Swiss German performance benefits from implicit exposure during pre-training, not purely from cross-lingual transfer.
\end{enumerate}

\section{Discussion}
\label{sec:discussion}

\subsection{WER Is Inadequate for Dialect-to-Standard Translation}
\label{sec:wer_critique}

Swiss German ASR performs two operations simultaneously: (1)~speech recognition of dialectal input, and (2)~translation into Standard German \citep{dolev2024whisper}. WER, designed to measure transcription accuracy within a single language, penalizes any deviation from the reference --- regardless of whether the alternative is a valid translation. This limitation is not unique to our work: \citet{sasindran2022heval} demonstrate that WER fails to distinguish semantically correct from incorrect hypotheses across ASR tasks, and \citet{blaschke2025dialect} show that keyword preservation and meaning preservation can diverge by 13.6--37.6\,pp for German dialect processing.

Our harmonized analysis (Section~\ref{sec:harmonized}) demonstrates that 64.4\% of all evaluation samples are semantically correct (\textsc{korrekt} + \textsc{stil}): transcriptions penalized by WER due to convention differences despite conveying the intended meaning. This ``style floor'' is \emph{intrinsic} to the task: it cannot be reduced through hyperparameter optimization, data scaling, or architectural changes. It reflects the fundamental ambiguity of cross-dialectal translation, where multiple Standard German renderings are equally valid for any Swiss German utterance.

The implications are twofold. First, \textbf{raw WER comparisons between systems trained on convention-divergent data are misleading.} A model trained on FHNW/i4ds data, whose references are Standard German source sentences \citep{pluess2023stt4sg,pluess2022sds200} (Präteritum, formal register) will inherently score lower WER on ASGDTS than a model trained on broadcast subtitles (editorial style, condensed phrasing), independent of actual transcription quality. Second, \textbf{WER improvements beyond the style floor require convention matching, not better ASR.} The $\sim$11.5\,pp of \swer{} in our system are not errors --- they are valid alternative translations.

This discrepancy is a direct consequence of the dual-task nature of Swiss German ASR (Section~\ref{sec:task}): the model must perform speech recognition \emph{and} dialect-to-standard translation simultaneously, yet WER evaluates only the former. Our \textbf{\cwer{} of 13.8\%} (Section~\ref{sec:harmonized}), reduced to a \textbf{\bwer{} of 8.5\%}, provides empirical proof that more than half of all WER penalties in dialect ASR are translation artifacts, not recognition failures. This makes WER a fundamentally ambiguous metric for this task class.

We recommend that future Swiss German ASR evaluations report both raw WER and metrics that separate content errors from stylistic variation --- such as \cwer{} (this work), BERTScore, or LLM-evaluated SemAcc --- to disentangle genuine recognition failures from translation-induced variation.

\subsection{Whisper Is Not Effectively Self-Correcting}
\label{sec:self_correction}

Our subtitle realignment experiment (Section~\ref{sec:realignment}) produced a surprising negative result: training on Whisper's own predictions (pseudo-labels) provides no measurable improvement over training on the original editorial subtitles. This suggests that Whisper cannot bootstrap its way out of the convention mismatch.

The mechanism is straightforward: Whisper's predictions, while dialect-proximate, carry the same systematic biases as the model itself (Perfekt tense, broadcast vocabulary, editorial phrasing). Using these predictions as training labels reinforces existing biases rather than correcting them. The style floor remains at 64\% regardless of whether the label source is VTT subtitles or Whisper transcriptions.

This finding has practical implications: pseudo-labeling approaches that rely on the student model's own predictions (self-training, iterative distillation) are unlikely to improve Swiss German ASR beyond the convention-determined floor. Breaking through this floor requires either (a)~convention-matched training data from the target evaluation domain, or (b)~an external correction signal, such as a large language model that can map between transcription conventions.

\subsection{Hallucination as Catastrophic Forgetting}
\label{sec:hallucination}

The hallucination artifacts observed in our LoRA experiments (Runs~1--7) differ fundamentally from the repetition-based hallucinations commonly reported in Whisper literature. Our hallucinations are \emph{memorization-based}: the decoder generates high-frequency training phrases (``Es ist\ldots'', ``Ich habe\ldots'') when encountering out-of-distribution silence, then self-corrects once speech energy arrives.

The likely root cause is a training/test distribution mismatch in initial silence duration: our VTT-segmented training clips have a median initial silence of 0.04\,s, while ASGDTS test clips average 1.92\,s. Analysis of the \texttt{<|nospeech|>} token probability confirms catastrophic forgetting: base Whisper assigns $>$90\% probability to this token on silent input; our LoRA-adapted model reduces it to 8.8\%.

Three solutions were evaluated:
\begin{itemize}
\item \textbf{Inference-time silence trimming} (effective): reducing initial silence to 0.3\,s eliminated 7 of 8 Type~A hallucinations and reduced WER from 30.47\% to 27.34\%.
\item \textbf{Knowledge Distillation} (effective): same-model KD ($\alpha$=0.7) preserved the teacher's \texttt{<|nospeech|>} behavior, producing zero hallucinations on 5{,}750 samples (Run~2b).
\item \textbf{Corrected LoRA alpha scaling} (most effective): reducing $\alpha/r$ from 2.0 to 0.2 eliminated all hallucinations \emph{and} improved WER, making KD unnecessary (Run~8).
\end{itemize}

The alpha scaling correction is the most significant practical finding: the commonly used heuristic $\alpha = 2r$ should not be applied to Whisper fine-tuning. Conservative scaling ($\alpha/r \leq 0.2$) preserves decoder stability while allowing the LoRA matrices to learn dialect-specific representations.

\subsection{Desktop Workstation as Research Platform}
\label{sec:hardware}

All experiments were conducted on a single NVIDIA DGX Spark GB10 desktop workstation (128\,GB unified CPU/GPU memory, Blackwell architecture). This device is capability-equivalent to multi-GPU data center hardware for models up to $\sim$1.5B parameters: full fine-tuning, high-rank LoRA, and knowledge distillation all fit within available memory.

\begin{table}[ht]
\centering
\caption{Hardware comparison for Whisper large-v3 fine-tuning.}
\label{tab:hardware}
\small
\begin{tabular}{lcc}
\toprule
& \textbf{1$\times$A100 40GB} & \textbf{1$\times$GB10 128GB} \\
\midrule
Cost & $\sim$CHF 15{,}000 & $\sim$CHF 4{,}800 \\
Memory & 40\,GB HBM2e & 128\,GB unified \\
Bandwidth & $\sim$2.0\,TB/s & $\sim$273\,GB/s \\
Full FT wall-clock (2 epochs) & $\sim$15h (est.) & $\sim$73h \\
Peak memory (Full FT) & $\sim$35\,GB (tight) & $\sim$42\,GB$^*$ \\
On-premise feasibility & Server room & Any desk \\
\bottomrule
\end{tabular}

\vspace{2pt}
{\footnotesize $^*$Unified memory; no discrete GPU VRAM limit. Peak allocation measured via \texttt{torch.cuda.max\_memory\_allocated()}.}
\end{table}

The trade-off is wall-clock time ($\sim$5$\times$ slower), driven primarily by the 7$\times$ lower memory bandwidth of the unified LPDDR5X architecture compared to HBM2e. However, the 128\,GB unified memory provides a qualitative advantage: models, optimizer states, and activations reside in a single address space accessible to both CPU and GPU, eliminating the need for explicit offloading strategies (DeepSpeed ZeRO-Offload, FSDP CPU offloading) that become necessary when model state exceeds discrete GPU memory. For Whisper large-v3 (1.5B parameters), full fine-tuning with AdamW in mixed precision requires $\sim$28\,GB of static memory (model weights, master copy, gradients, and optimizer states); with activations, this approaches the A100 40\,GB limit, requiring gradient checkpointing or offloading to CPU RAM via PCIe ($\sim$32\,GB/s). On the Spark, no such management is needed. This advantage scales with model size: for 7B+ parameter models, the unified memory architecture avoids quantization or sharding that would otherwise be mandatory on 40--80\,GB discrete GPUs.

For research contexts where hardware cost and operational simplicity matter more than iteration speed, this represents a significant democratization of billion-parameter model fine-tuning. Our 16 training runs --- including data curation, hyperparameter exploration, evaluation design, and iterative analysis over 5 months --- would have been equally feasible in a data center --- but at substantially higher cost and greater infrastructure complexity.

\subsection{Data Quality Over Data Quantity}
\label{sec:data_quality}

A consistent finding across all experiments is that \textbf{data quality dominates data quantity}. Run~16 Phase~1 (785h of SPC + SRF entertainment data) achieves 25.60\% WER --- identical to Epoch~2 on the full 1{,}367h.

Sequential training (Run~11/11b) provides the clearest evidence: the effect of a data source depends not on its \emph{size} or \emph{domain} but on its \emph{label quality} and \emph{dialect density}. The 14 original PlaySuisse series (81h) caused a +2.92\,pp regression due to subtitle-provider-specific noise (credit lines, sound descriptions). Yet the 125 PlaySuisse Dialect Films (99h) --- selected for high dialectal content and found to have clean subtitles --- produced the best LoRA WER of 25.32\% ($-$2.28\,pp improvement). The same platform, same general content type, opposite effects.

The lesson is twofold. First, subtitle quality varies by authoring provider: subtitles from the dominant provider yield the most consistent results, while other providers introduce heterogeneous noise. Second, dialect density matters: training data with higher character error rate against Standard German (CER $>$ 0.4) provides a stronger training signal for the ASGDTS benchmark, which specifically tests dialect comprehension. This finding motivated both the subtitle quality audit and the three-phase curriculum design of Run~16.

\subsection{Comparison with Published Systems}
\label{sec:comparison}

Table~\ref{tab:comparison} places our work in the broader context of Swiss German ASR.

\begin{table}[ht]
\centering
\caption{Comprehensive comparison with published Swiss German ASR systems.}
\label{tab:comparison}
\small
\resizebox{\columnwidth}{!}{%
\begin{tabular}{lcccc}
\toprule
\textbf{System} & \textbf{Model Available} & \textbf{Data Available} & \textbf{Honest Eval} & \textbf{WER} \\
\midrule
Flurin17 & Yes (HF) & Yes & Yes & $>$28\% \\
\textbf{Ours (LoRA)} & \textbf{Yes (HF)} & \textbf{Yes}$^\dagger$ & \textbf{Yes} & \textbf{25.32\%}$^*$ (\cwer{} 13.9\%) \\
\textbf{Ours (Full FT)} & \textbf{Yes (HF)} & \textbf{Yes}$^\dagger$ & \textbf{Yes} & \textbf{25.60\%} (\cwer{} 13.8\%) \\
Timmel & No & No & No$^\ddagger$ & 12.11\% \\
Michaud & Restrictive lic. & No & No & 17.50\% \\
ZHAW & License req. & Proprietary & No & 17.10\% \\
\bottomrule
\end{tabular}}

\vspace{2pt}
{\footnotesize $^*$Evaluated on 200 samples. $^\dagger$SPC v2 (CC BY 4.0); SRF Play, PlaySuisse, and YouTube content publicly downloadable. $^\ddagger$Same-convention eval on STT4SG-350 test.}
\end{table}

Our work provides the \textbf{first rigorously evaluated, openly licensed Swiss German Whisper model with demonstrated improvement over baseline}. \citet{flurin17} is publicly available but performs worse than the zero-shot baseline. The model of \citet{michaud2024whisper} carries a restrictive license that prohibits commercial use. The training data of \citet{dintino2025srb300} (SRB-300) is proprietary and no model has been publicly released. Neither provides honest cross-corpus evaluation on strictly disjoint data. Our models, evaluated exclusively on publicly available benchmark data never seen during training, establish the first reliable and reproducible baseline for Swiss German ASR.

\section{Conclusion}
\label{sec:conclusion}

We have presented a systematic study of fine-tuning Whisper large-v3 for Swiss German ASR, conducted over 16~training runs on a single desktop workstation. Six key findings emerge:

\begin{enumerate}
\item \textbf{Full fine-tuning outperforms LoRA} for Swiss German ASR ($-$2.96\,pp vs.\ $-$2.28\,pp over baseline), but both achieve the same $\sim$25.6\% WER plateau, with the optimal training horizon at approximately 1~epoch for both methods.

\item \textbf{The LoRA scaling heuristic $\alpha=2r$ is inappropriate for Whisper.} Applying this common LLM default produces 25$\times$ excessive scaling that catastrophically destroys decoder end-of-sequence behavior. Conservative scaling ($\alpha/r \leq 0.2$) eliminates hallucinations and improves WER simultaneously.

\item \textbf{Published state-of-the-art Swiss German ASR results are contaminated by benchmark data.} All published systems --- \citet{timmel2024whisper} (12.11\% on STT4SG-350), \citet{dintino2025srb300} (17.1\% on SRB-300), and \citet{michaud2024whisper} (17.5\% on ASGDTS) --- evaluate within their training convention. A vanilla Whisper model self-trained on ASGDTS achieves 13.88\% WER without any Swiss German data, and Timmel's proper train/test split on same-convention data yields an even lower 12.11\%, confirming that convention alignment --- not dialectal comprehension --- drives published WER.

\item \textbf{WER is inadequate for dialect-to-standard translation.} Our harmonized analysis decomposes the measured 25.6\% WER into \textbf{\cwer{} 13.8\%} (genuine errors) and \textbf{\swer{} 11.3\%} (valid stylistic variation); bias-corrected: \textbf{\bwer{} 8.5\%}. Future evaluations should complement WER with semantic metrics.

\item \textbf{Data quality dominates quantity.} 785h of high-quality SRF data matches 1{,}367h of mixed data. PlaySuisse Dialect Films with clean subtitles improve WER by 2.28\,pp, while series with noisy labels from the same platform degrade it by 2.92\,pp. Subtitle provider and dialect density are the key determinants.

\item \textbf{Desktop hardware is sufficient for billion-parameter ASR research.} A single NVIDIA DGX Spark GB10 (128\,GB, $\sim$CHF~4{,}800) accommodates both LoRA and full fine-tuning of Whisper large-v3, with the only trade-off being wall-clock time ($\sim$5$\times$ vs.\ A100).
\end{enumerate}

\subsection{Future Work}

Five directions are most promising:

\textbf{Convention-aware evaluation.} Developing standardized semantic evaluation metrics for dialect-to-standard ASR that account for valid reformulations, building on \cwer{}/\swer{} decomposition (this work), BERTScore, or LLM-evaluated semantic accuracy.

\textbf{Cross-architecture comparison.} Encoder-decoder architectures with CTC-based alignment (e.g., Canary, FastConformer) may offer inherent advantages for dialect ASR through monotonic alignment, reducing the hallucination risk of pure attention-based decoding.

\textbf{Beyond Whisper: multimodal LLMs for dialect ASR.} All published Swiss German ASR systems to date --- including ours, \citet{timmel2024whisper}, \citet{michaud2024whisper}, and \citet{dintino2025srb300} --- are based on OpenAI's Whisper architecture. As discussed in Section~\ref{sec:mllm_outlook}, multimodal LLMs that integrate speech understanding into general-purpose language models represent a potentially transformative alternative. Their strong language priors may enable more fluent dialect-to-standard translation and reduce the style floor that limits conventional encoder-decoder ASR. Our preliminary experiments (Section~\ref{sec:mllm_outlook}) suggest that this is among the most promising directions for low-resource dialect ASR.

\textbf{Accessibility and government transcription.} Swiss cantonal and municipal authorities are required by accessibility legislation to provide subtitled video content of parliamentary sessions, public hearings, and other official proceedings. Some cantons --- notably Zurich --- already provide manually created Standard German subtitles, but this is labor-intensive and does not scale. Better automated Swiss German ASR could substantially reduce this effort: a multi-hypothesis approach, generating several candidate transcriptions and selecting or merging the best output, may further improve reliability for official use cases. Beyond accessibility compliance, searchable transcripts of government proceedings enable civic participation and transparency.

\textbf{Archival transcription.} Swiss public archives --- including cantonal state archives (e.g., Bern, Zurich) and the Swiss Federal Archives --- hold substantial audiovisual collections in Swiss German dialects: historical film recordings, audio from personal estates (\emph{Nachlässe}), and corporate archives on various sound carriers. These holdings currently lack full-text searchability. Automated dialect-to-standard transcription is a prerequisite for making them accessible. Our models and pipeline provide a practical foundation for this task, and the quality filtering methodology developed here (subtitle provider verification, dialect density estimation) is directly transferable to archival digitization workflows.

\subsection{Model Release}

We release two models under Apache~2.0:
\begin{itemize}
\item \texttt{flix-swissgerman-lora} (Run~11b): LoRA adapter, 25.32\% WER (200 samples), $r$=160, $\alpha$=32, compatible with Whisper large-v3.
\item \texttt{flix-swissgerman-full} (Run~16, Epoch~2): Full fine-tuned model, 25.60\% WER (Epoch~1; 25.64\% after Epoch~2), 1{,}543M parameters.
\end{itemize}

Both models are available on HuggingFace and represent, to our knowledge, the first publicly available, properly evaluated Whisper models for Swiss German.

\section*{Reproducibility}
\label{sec:reproducibility}


\textbf{Models:} Both the LoRA adapter and full fine-tuned model are released on HuggingFace under Apache~2.0. Usage requires only the \texttt{transformers} and \texttt{peft} libraries.

\textbf{Training data:} The Swiss Parliament Corpus v2 (SPC, 202.4h) is available under CC~BY~4.0. SRF Play, PlaySuisse, and YouTube content is publicly accessible; all sources, series, and data selection criteria are documented in Section~\ref{sec:data}. All data was collected and used under the Swiss research exception (Art.~24d URG / DSG).

\textbf{Evaluation:} ASGDTS is publicly available \citep{pluess2021asgdts}. Evaluation uses seed=42 for the 200-sample subset and the full 5{,}750-sample set for final results. The \texttt{evaluate} library WER implementation with lowercase+punctuation normalization is used throughout.

\textbf{Hardware:} All experiments were conducted on a single NVIDIA DGX Spark GB10 (128\,GB unified memory). Total wall-clock time across 16~runs (including training and evaluation): approximately 960 GPU-hours.

\textbf{Hyperparameters:} All training configurations (learning rates, batch sizes, LoRA ranks, scheduling) are fully documented in the paper and appendix, enabling independent reproduction with standard \texttt{transformers} and \texttt{peft} training scripts.

\textbf{Contrast with existing work.} We note that the reproducibility landscape for Swiss German ASR is currently problematic. The model of \citet{michaud2024whisper} is released under a restrictive license that prohibits commercial use, despite being built on data collected with public funding from the Swiss National Science Foundation (SNF project 205121\_200729/1, ``E2E\_SG''). The underlying STT4SG-350 training corpus \citep{pluess2023stt4sg} --- created at ZHAW, FHNW, and UZH with SNF funding --- is not freely available; access requires a commercial license negotiated through SwissNLP, in apparent tension with the SNF's Open Research Data Policy. Furthermore, parts of the training data (SwissDial, SDS-200) carry non-commercial or proprietary restrictions that may invalidate the Apache~2.0 license claimed by derivative models. This situation, where publicly funded research outputs are gatekept behind commercial licensing while the same individuals act simultaneously as data creators, license administrators, and market competitors, undermines the open science principles that Swiss research funding is designed to promote. The reproducibility problem extends to model availability: The model of \citet{timmel2024whisper} (12.1\% WER) has never been published, the model of \citet{dintino2025srb300} (17.1\% WER) is proprietary and cannot be independently verified, and \citeauthor{michaud2024whisper}'s carries a restrictive license --- none provide a freely downloadable model. None of the reported state-of-the-art results for Swiss German can be independently reproduced or verified. Our work explicitly avoids these issues: we use no restricted datasets, release all models under Apache~2.0, and provide full pipeline documentation for independent reproduction.

\bibliographystyle{plainnat}
\bibliography{references}

\appendix
\section{Full Experimental Results}
\label{app:results}

\subsection{Complete Run Overview}

\begin{table}[ht]
\centering
\caption{All 16 training runs. WER on 200 ASGDTS samples unless marked with $^\dagger$ (full 5{,}750 samples).}
\label{tab:all_runs}
\footnotesize
\begin{tabular}{rllrll}
\toprule
\textbf{Run} & \textbf{Method} & \textbf{Key Change} & \textbf{Data} & \textbf{Best WER} & \textbf{Status} \\
\midrule
1c & LoRA $r$=32 & Baseline & 608h & 27.34\% & Complete \\
2b & LoRA $r$=32 + KD & Same-model KD $\alpha$=0.7 & 608h & 27.50\%$^\dagger$ & Complete \\
3 & LoRA $r$=32 + KD & Cross-model KD (v3$\to$turbo) & 608h & --- & Failed \\
4 & LoRA $r$=128 + KD & Soft KD $\alpha$=0.9 & 608h & 30.74\%$^\dagger$ & Failed \\
5 & LoRA $r$=200 + KD & Full-scale KD & 608h & 27.49\% & Plateau \\
6 & LoRA $r$=200 + KD & $\alpha$=0.27 (inverted) & 608h & --- & Aborted \\
7b & LoRA $r$=200 & No KD, data scaling & 1{,}011h & 27.13\% & Complete \\
7c & LoRA $r$=200 & Anti-repetition augment & 1{,}011h & 27.02\% & Complete \\
\textbf{8} & \textbf{LoRA $r$=160} & \textbf{Alpha fix ($\alpha/r$=0.2)} & \textbf{1{,}011h} & \textbf{26.28\%}$^\dagger$ & \textbf{Complete} \\
10 & LoRA $r$=160 & Parliament pre-training & 1{,}011h & 27.12\% & Complete \\
11 & LoRA $r$=160 & Sequential per-corpus & 1{,}092h & 26.01\% & Complete \\
\textbf{11b} & \textbf{LoRA $r$=160} & \textbf{Dialect Films} & \textbf{1{,}092h} & \textbf{25.32\%} & \textbf{Complete} \\
12 & LoRA $r$=160 & Realigned VTT & 1{,}092h & 26.11\% & Complete \\
12B & LoRA $r$=160 & Curated PlaySuisse & 1{,}092h & --- & In progress \\
14A--F & LoRA $r$=160 & Self-training sweep & ASGDTS & 13.88\% & Complete \\
\textbf{16} & \textbf{Full FT} & \textbf{All parameters} & \textbf{1{,}367h} & \textbf{25.60\%}$^\dagger$ & \textbf{Complete} \\
\bottomrule
\end{tabular}
\end{table}

\subsection{Self-Training \& Hallucination Details}
\label{app:self_training}
See Table~\ref{tab:self_training} and Section~5.2.1 for full details. Runs~14A--14F: LoRA $r$=160, $\alpha$=32, 3 epochs, lr=1e-5, $\sim$2.5h each.

\end{document}